\documentclass[10pt,journal,twocolumn,twoside]{IEEEtran} 
\usepackage{graphicx}
\usepackage{subcaption}
\usepackage{epstopdf}
\usepackage{float}
\usepackage{algorithmic}
\usepackage{array}
\usepackage{amsmath}
\usepackage{amssymb}
\usepackage{mdwmath}
\usepackage{mdwtab}
\usepackage{eqparbox}
\usepackage{stfloats}
\usepackage{fixltx2e}
\usepackage{hyperref}
\usepackage{cases} 
\usepackage{flushend}
\usepackage{upgreek}
\usepackage{booktabs}
\usepackage{multirow}
\usepackage{makecell}
\usepackage{xcolor}
\usepackage{makecell}
\usepackage{url}
\usepackage{cite}
\usepackage{algorithm}  

\usepackage{textcomp}
\usepackage{bm}
\usepackage{setspace}
\usepackage{epsfig}
\usepackage{diagbox}
\allowdisplaybreaks[4]

\makeatletter
\renewcommand*{\@opargbegintheorem}[3]{\trivlist
      \item[\hskip \labelsep{\bfseries #1\ #2}] \textbf{(#3):}\ }
\makeatother     

\begin{document}

\title{Large Language Model as a Catalyst:  A Paradigm Shift in Base Station Siting Optimization}

\author{ Yanhu Wang,  Muhammad Muzammil Afzal,  Zhengyang Li, Jie Zhou, Chenyuan Feng, {\it {Member, IEEE}}, \\   Shuaishuai Guo, {\it {Senior Member, IEEE}}, and Tony Q. S. Quek, {\it {Fellow, IEEE}}
\thanks{Yanhu Wang, Muhammad Muzammil Afzal, Zhengyang Li, Jie Zhou, and Shuaishuai Guo are with School of Control Science and Engineering, Shandong University, Jinan 250062, China (e-mail: \{yh-wang,zhengyang\_li,jiezhou\}@mail.sdu.edu.cn; muzammil\_afzaal@yahoo.com; shuaishuai\_guo @sdu.edu.cn) } 
\thanks{Chenyuan Feng is with the Department of Communication Systems, EURECOM, Sophia Antipolis 06410, France  (e-mail: Chenyuan.Feng@eurecom.fr).}
\thanks{Tony Q. S. Quek is with the Information Systems Technology and Design Pillar, Singapore University of Technology and Design, Singapore 487372 (e-mail: tonyquek@sutd.edu.sg).} }

\pagestyle{empty}
\maketitle

\begin{abstract}
Traditional base station siting (BSS) methods rely heavily on drive testing and user feedback, which are laborious and require extensive expertise in communication, networking, and optimization. As large language models (LLMs) and their associated technologies advance, particularly in the realms of prompt engineering and agent engineering, network optimization will witness a revolutionary approach. This approach entails the strategic use of well-crafted prompts to infuse human experience and knowledge into these sophisticated LLMs, and the deployment of autonomous agents as a communication bridge to seamlessly connect the machine language based LLMs with human users using natural language. Furthermore, our proposed framework incorporates retrieval-augmented generation (RAG) to enhance the system's ability to acquire domain-specific knowledge and generate solutions, thereby enabling the customization and optimization of the BSS process. This integration represents the future paradigm of artificial intelligence (AI) as a service and AI for more ease. This research first develops a novel LLM-empowered BSS optimization framework, and heuristically proposes three different potential implementations: the strategies based on Prompt-optimized LLM (PoL), LLM-empowered autonomous BSS agent (LaBa), and Cooperative multiple LLM-based autonomous BSS agents (CLaBa).  Through evaluation on  real-world data, the experiments demonstrate that prompt-assisted LLMs and LLM-based agents can generate more efficient and reliable network deployments, noticeably enhancing the efficiency of BSS optimization and reducing trivial manual participation. 
\end{abstract}


\IEEEpeerreviewmaketitle

\thispagestyle{empty}

\section{Introduction}

\IEEEPARstart{A}{s} the backbone of mobile communication networks, base stations play a pivotal role  in delivering uninterrupted connectivity to mobile users and also catering to the escalating appetite for high data throughput, ensuring the seamlessness and dependability of communications \cite{ref001,10015950}.  This capability empowers users to relish high-speed network services, irrespective of their mobility. The process of identifying the most advantageous positions for base station installations within a communication network, such as those for 4G/5G cellular networks, is known as base station siting (BSS). The overarching objective is to amplify network coverage, signal excellence, and network capacity, while concurrently curbing deployment expenses and mitigating environmental footprints.  With the proliferation of smartphones and mobile devices, there has been a meteoric rise in the number of mobile users, alongside a proportional increase in the demand for swift and superior data quality \cite{10293197,10458014, 7997740}. Consequently, on-demand BSS has become exceedingly critical and challenging, as it exerts a profound influence on the expanse and intensity of network coverage and the quality of service (QoS) experienced by users \cite{6489498,wu2015energy,10086045}.

\subsection{Related Works of Traditional Methods}
Conventional BSS techniques primarily rely on road testing and user feedback to evaluate network performance and identify areas for improvement \cite{ref002,ref003}. This process requires communications engineers to undertake several key steps to ensure that new base stations are effectively deployed to enhance network coverage and user experience.

First, conduct road tests by driving test vehicles through urban areas to measure and record signal strength, coverage, and data transmission rates. This provides engineers with  a dispassionate assessment of current network performance and help identify weak coverage zones and blind spots.  Subsequently, user feedback is gathered, typically through customer service channels or mobile apps, where users report issues such as dropped calls, weak signals, or unstable data connections.  The engineer then compiles all of this user feedback into a thorough problem report. Engineers compile this feedback into a comprehensive report, which complements the road test data by highlighting additional issues related to the actual user experience. After gathering sufficient information, engineers model potential base station locations, considering factors such as topography, building obstructions, subscriber density, and the configuration of existing base stations.

The next critical step is solution development, where engineers determine the optimal location for the new base station based on the collected data and also the model analysis. To achieve this, optimization algorithms such as simulated annealing \cite{10070298}, genetic algorithm (GA) \cite{li20236g}, or particle swarm optimization (PSO) \cite{li2022particle} are used to balance factors like coverage effectiveness, construction costs, and operational efficiency. Once the optimal location is identified, the base station is deployed, and its performance is monitored through further road tests and user feedback. Engineers may need to make additional adjustments to ensure the new base station is functioning at its full potential.

While effective, this conventional approach has several limitations. Road testing, though informative, is time-consuming and logistically challenging, especially in densely populated urban areas \cite{tayal2020optimization}. Moreover, the data collected during road testing represents only a specific moment in time and location, which may not capture the dynamic variations in network performance over time \cite{malekzadeh2023performance}. User feedback, while valuable, is often reactive, meaning that network improvements are usually initiated only after issues have become severe enough to prompt complaints. Furthermore, the feedback may not fully represent the broader user base, potentially leading to biased or incomplete data \cite{9355403,6787081}. Engineers are thus required to engage in an ongoing, iterative process of feedback analysis, problem modeling, solution development, base station deployment, and network performance reevaluation \cite{electronics8111318}.

Given these challenges, traditional BSS methods demand a high level of expertise in communications, networking, optimization, and programming. Engineers must also possess strong analytical and problem-solving skills to navigate the increasing complexity of the task. The rapid pace of advancements in telecommunications technology \cite{8382166,9144301} and shifting user behavior patterns \cite{9720630} require engineers to continuously learn and adapt. Additionally, the dynamic nature of urban environments—characterized by fluctuating traffic patterns \cite{9678040}, user mobility \cite{9759241}, and varying service demands over time \cite{10489248}—further complicates the process of BSS optimization.

\subsection{Motivations for Incorporating LLMs}

The integration of AI, particularly LLMs, offers unprecedented potential to navigate the escalating intricacy and dynamism inherent in next-generation networks. Models such as Generative Pretrained Transformers (GPT)-3.5, GPT-4, and GPT-4o have emerged as paragons of advanced natural language processing (NLP) prowess. These sophisticated models are capable of producing text that closely mimics human beings \cite{10177738, chang2023learning} and are adept at resolving multifaceted challenges spanning a spectrum of disciplines, including mathematics \cite{didolkar2024metacognitive}, programming \cite{nijkampcodegen}, and computer vision \cite{hu2024bliva}. The advent of these models empowers users to express their specifications in natural language, thereby catalyzing a pivotal transition from semi-automated to fully automated modeling and coding paradigms \cite{charlie2024}. This paradigm shift liberates professionals to concentrate on nuanced problem-solving and pioneering design endeavors.
For instance, LLMs, combined with the mixture of human experts, significantly optimized the transmission strategy of the satellite network in \cite{10679152}. \cite{li2024large} proposed to use LLMs to solve the multi-objective optimization problem in integrated sensing and communication (ISAC) systems. In mobile networks, LLMs automate the course design of reinforcement learning, thereby improving the convergence speed and performance of learning agents \cite{10682015}. In the vehicular networks, \cite{10683673} used LLMs to optimize resource allocation between vehicles and roadside units, which greatly improves the efficiency and performance of the system. Overall, this integration of AI into the fabric of network optimization not only streamlines existing processes but also paves the way for groundbreaking innovations in the field.

Regarding the BSS problem, LLMs can streamline the network optimization process through prompt engineering, where complex communication issues are translated into structured tasks. By carefully designing prompts, LLMs can understand and generate optimization strategies, allowing engineers to rapidly develop effective solutions \cite{telecompgt,zhou2024}. This approach not only enhances the efficiency of problem-solving but also reduces the need for manual analysis and intervention. Additionally, leveraging AI agent engineering enables LLMs to function as intelligent agents within communication networks. These agents can continuously monitor network conditions, process user feedback, automatically adjust network parameters, and make optimization decisions in real-time, significantly alleviating the workload of network engineers. LLMs, with their robust reasoning and learning capabilities, can respond to dynamic network environments and provide optimal decisions promptly. This automation enhances operational efficiency, reduces response times, optimizes resource allocation, and improves overall user experience.

Additionally, LLMs offer several notable benefits:  i) They can process vast amounts of real-time data from various sources by employing open-source algorithms tailored to specific sub-problems, resulting in a comprehensive analysis of network performance. This capability enables more efficient and accurate identification of weak coverage areas and service deficiencies. ii) LLMs can reduce the delays associated with passive feedback mechanisms by proactively suggesting improvements based on continuous learning from network data and user feedback.  iii) Their dynamic adaptability to changes in traffic patterns and user behavior ensures that the generated base station solutions remain relevant and effective in rapidly evolving urban environments.

Therefore, the introduction of LLMs not only elevates the level of intelligence in BSS optimization but also, through prompt engineering and AI agent engineering, empowers engineers to tackle complex issues more efficiently and autonomously, driving advancements in future network optimization technology.

\subsection{ Contributions }
In response to the burgeoning potential of LLMs in the realm of communication networks, this research investigates how LLMs may revolutionize BSS by enhancing both the efficacy of the siting process and the overall quality of mobile network services.  Specifically, we propose an innovative LLM-empowered paradigm for BSS problem, characterized by three distinct strategies that are delineated based on the level of human involvement and the interplay between autonomous agents, namely, Prompt-optimized LLM-based (PoL-) strategy, LLM-empowered autonomous BSS agent-based (LaBa-) strategy, and Cooperative multiple LLM-based autonomous BSS agents-based (CLaBa-) strategy. Additionally, our framework integrates retrieval-augmented generation (RAG), enabling the system to dynamically extract precise expert knowledge from external sources and adaptively learn this information to enhance the BSS process.

To the best of our knowledge, we are the first to explore the use of LLM and RAG to solve the BSS problem. Our main contributions in this work can be summarized as follows:
\begin{itemize}
    \item Framework Formulation and Strategy Design: We formulate a pioneering LLM-empowered BSS paradigm, supported by three distinct, heuristically designed strategies. Specifically, the PoL strategy facilitates autonomous LLM execution of BSS tasks with minimal human intervention; the LaBa strategy propels the vision of a fully independent, end-to-end BSS process; and the CLaBa strategy is meticulously crafted to further improve system efficiency, enhance robustness, and address complex problems. By incorporating RAG, we further enhance these strategies by allowing LLMs to access real-time, contextually relevant information from large knowledge bases, thus improving the precision and adaptability of BSS solutions. 
    \item  Experiment Simulation and Analysis: We execute an empirical comparative analysis by leveraging  a real-world dataset \cite{mathorcup2022}. This analysis thoroughly evaluates the performance of LLM-driven approaches, focusing on metrics such as traffic coverage and cost-effectiveness. Moreover, given RAG's powerful ability of retrieving domain-specific knowledge, the experiments demonstrate that the RAG-enhanced LLM strategies outperform baseline models in terms of solution accuracy and robustness. The results provide compelling evidence for the practical viability of LLM and RAG-powered strategies in real-world applications.
\end{itemize}

Besides propelling technological advancements in BSS, we also introduce fresh perspectives and tools to the telecom network design domain. Through these innovative approaches, we are able to achieve more efficient and reliable network deployment to fulfill the expanding demand for communications, while optimizing resource allocation and reducing operational costs. {Furthermore, our research delineates several frameworks intended to guide future investigators in their quest to refine and innovate paradigms for harnessing LLMs and RAG to resolve intricate engineering challenges.}

The rest of this paper is structured as follows. Section II introduces the BSS problem, providing a detailed description. The strategies based on prompt engineering and agent engineering are demonstrated in Sections III. Section IV integrates RAG technology into the proposed strategies for enhanced performance. Section V presents the numerical results. Section VI deeply discusses the future research direction, and lastly, Section VII concludes this paper. 

\newcolumntype{M}[1]{>{\centering\arraybackslash}m{#1}}

\section{Problem Description}

\begin{figure}
    \centering
    \includegraphics[width=0.45\textwidth]{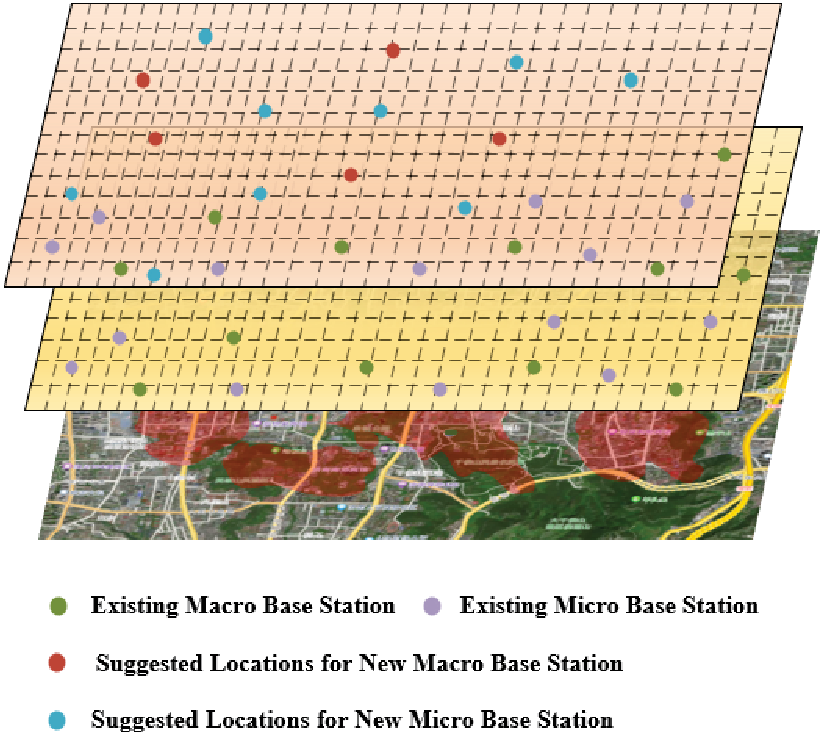}
    \caption{\textbf{The coverage and planning of base stations within a given region.} The real-world map is shown on the bottom layer; existing macro and micro base stations are displayed on the middle layer; both planned and existing macro and micro base stations are marked in the top layer, along with proposed upgrades to address areas with poor coverage.}
    \label{fig1}
\end{figure}

\subsection{Network Model}
In this work, we consider a BBS problem in a real-world communication network, as shown in Fig. \ref{fig1}. Specifically, the bottom layer represents the communication coverage of the existing network across a specified urban landscape, with red zones highlighting areas where communication coverage is sub-optimal. Identifying such areas can be achieved through costly and laborious road tests or by analyzing user feedback regarding signal quality.

By dividing complex geographical areas into smaller grids, the BBS selection problem can be significantly simplified, making it more manageable for mathematical modeling and algorithmic approaches. This method reduces the total number of potential site candidates by focusing solely on grid centers, which not only minimizes computational demands but also improves the overall efficiency of the site selection process. Furthermore, a grid-based approach ensures a more uniform coverage of the region, preventing the risks of either leaving certain areas underserved or oversaturating others with base stations. Therefore, in this work, we divide the target area into multiple grids and consider only the central points of each grid as candidate sites for new base stations, as shown in the middle layer of Fig. \ref{fig1}. 

In the segmentation of the grid, we posit that the coverage radius of both macro and micro base stations is an integer multiple of the grid radius. This assumption ensures that a base station, once deployed at the centroid of one small grid, can offer communication coverage across the entire grid. Such partitioning guarantees that irrespective of the region's expanse, the candidate locations for new base stations can be represented as a finite set of points. The BSS decision-making process is thus anchored on the specific characteristics of each point, encompassing factors such as coordinates, the quality of communication coverage, and traffic volume.

\subsection{Problem Formulation}
\subsubsection{Objective}
The main goal of BSS is to pinpoint regions within the current network that suffer from inadequate coverage and to strategically situate new base stations to augment connectivity in these zones \cite{chen2023base}.  In the realm of pragmatic network planning, it is often impractical to address all coverage deficiencies at once, given the substantial financial outlay required for constructing base stations. Consequently, there is a pressing need to prioritize areas with weak coverage but high traffic density for targeted enhancement.

\subsubsection{Constraints}
When embarking on the deployment of a novel base station, the primary goal is to ensure the most extensive seamless coverage feasible, particularly in areas with significant traffic flow.  In addition, to mitigate interference and to consider deployment expenses, a minimum threshold distance must be maintained between any two stations.  Telecommunication operators are tasked with achieving a balance between reduced costs and fulfilling signal coverage mandates through a judicious deployment strategy that encompasses both macro and micro base stations. Macro base stations, known for their expansive coverage radius and higher construction costs, are ideal for broad area coverage. In contrast, micro base stations, with their lower costs and focused coverage, are optimally suited for augmenting network capacity and providing supplementary coverage in specific locations, such as traffic hotspots.

\subsection{Objective Function}
Taking into account the attributes of base stations, desired network coverage, and the financial implications of deployment, the optimization problem in BSS can be articulated as follows: 
\begin{equation}\label{eq:P1}
\begin{aligned}
& (\text{P1}) ~ {\arg\min}_{ \{ (p_i, q_i) \}_{i=1}^N}~ \sum_{i=1}^N \left( p_i C_h + q_i C_d \right) ,\\
s.t.
& (\text{C1})~ \sum_{ t \in G_i } w_{t} (P_{i,h,t} + P_{i,d,t}) \geq \theta_{cp} \sum_{t \in G_i} w_{t}, \forall i \in \mathcal{N}, \\ 
& (\text{C2})~ p_i \in \{0, 1\}, ~~ q_i \in \{0, 1\}, ~~\forall i \in \mathcal{N}, \\
& (\text{C3})~ p_i + q_i \leq 1, ~~\forall i \in \mathcal{N},\\
& (\text{C4})~ \sqrt{(x_i - x_j^e)^2 + (y_i - y_j^e)^2} \geq D_{\text{min}}, \\
& ~~~~~~~~\text{if $p_i+q_i=1$ },~ \forall i \in \mathcal{N}, \forall j \in  \mathcal{T}, \\
& (\text{C5})~  \sqrt{(x_i - x_n)^2 + (y_i - y_n)^2} \geq D_{\text{min}}, \\
& ~~~~~~~\text{if $p_i+q_i= 1$ and $p_n+q_n=1$},~ \forall i,n \in \mathcal{N},
\end{aligned} 
\end{equation}
with 
\begin{equation}
\begin{aligned}
P_{i,h,t} & =P\{p_i \sqrt{(x_i - x_t)^2 + (y_i - y_t)^2} \leq  d_h\},\\
P_{i,d,t} &=P\{q_i \sqrt{(x_i - x_t)^2 + (y_i - y_t)^2} \leq  d_d\},
\end{aligned}
\end{equation}
where $P_{i,h,t}$ and $P_{i,d,t}$ represent the probabilities that a device located at $t \in G_i$, with coordinates $(x_t,y_t)$, falls within the coverage of a macro base station or a micro base station situated at $(x_i,y_i)$, respectively; $G_i$ signifies the entire area encompassed by grid $i$; $w_t$ corresponds to the traffic volume at the location defined by $(x_t,y_t)$; { $\theta_{cp}$ denotes a predefined threshold of data traffic coverage probability}; the set $\mathcal{N}$ comprises the coordinates of all potential locations for new base stations, and $N$ is the aggregate number of grid points under consideration; the set $\mathcal{T}$ represents the coordinates of all current base stations in operation; the parameters $d_h$ and $d_d$ denote the coverage radii for macro and micro base stations, respectively, while $C_h$ and $C_d$ represent the respective setup costs for these stations; $D_{min}$ defines the minimum allowable distance between any two base stations to ensure effective interference mitigation and cost management; the Boolean variables $p_i$ and $q_i$ indicate the presence of a macro base station and a micro base station at the centroid of grid $i$, respectively; $(x_i, y_i)$ is the coordinates of the central point of grid $i$; and $(x^e_j, y^e_j)$ is the coordinates of an existing base station $j$.

In the optimization model P1, C1 stipulates that the probability of data traffic coverage must exceed the threshold $\theta_{cp}$, C2  dictates the binary choice of whether to deploy a base station at a candidate location; C3 prohibits the construction of more than one new base station at a single site, C4 mandates that the distance between any two new base stations be greater than $D_{\text{min}}$, C5 requires that new base stations be situated at least $D_{\text{min}}$ away from any existing stations. These constraints are designed to optimize the placement of base stations for maximum coverage while minimizing interference and deployment costs.

\begin{figure*}[ht!]
\centerline{\includegraphics[width=0.85\textwidth, keepaspectratio]{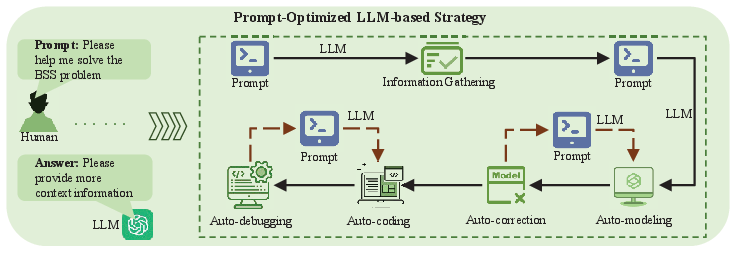}}
\caption{Diagram of the prompt-optimized LLM-based (PoL) strategy, showcasing the iterative workflow involving information gathering, automatic modeling \& optimization, as well as automatic code generation \&  correction (including auto-coding, auto-debugging, and also auto-correction). This workflow is guided by human-initialized prompts, enabling efficient solutions to the BSS problem.}
\label{fig2}
\end{figure*}

\section{LLM-empowered BSS Paradigm}
\subsection{Strategy based on Prompt Engineering}
{The LLM-aided BSS paradigm based on prompt engineering centers on the use of carefully crafted prompts to guide LLMs in generating desired outputs. Users or engineers provide specific inputs, and the model's response depends on the clarity and design of these prompts. The effectiveness of this method hinges on optimizing the prompt structure to elicit precise and relevant responses from the model. Human involvement is crucial throughout, as engineers must continuously adjust the prompts, interpret the generated results, and fine-tune based on feedback to achieve optimal outcomes. This workflow is inherently iterative, with frequent trial-and-error to improve performance, driven by user input. When done effectively, prompt engineering can produce high-quality, tailored solutions.  Once an engineer successfully designs effective prompts, they can be easily adapted for different tasks without requiring deep technical knowledge, allowing for broader applicability across domains.}

{The major challenge lies in creating prompts that are clear and precise, as this directly influences the model's accuracy and relevance. In this regard, we design a LLM-based strategy for BBS optimization based on prompt engineering: the Prompt-Optimized LLM-based (PoL-) strategy. Moving forward, we will delve into a detailed exposition of its workflow.}

\subsubsection{Workflow of PoL Strategy}
{The core of the PoL strategy is to guide the LLM to complete the BSS task automatically through well-designed prompts. By designing the right prompt, the LLM can understand the key requirements of the BSS problem and generate the appropriate optimization solution. However, this approach involves more than a simple ``Q \& A" model; it is a complex iterative process. Each prompt must be carefully designed to meet the specific needs of the BSS, ensuring that the LLM can accurately identify critical issues and propose effective solutions.}

{As shown in Fig. \ref{fig2}, the workflow of the PoL strategy can be divided into the following steps: \textbf{i) Information Gathering}: LLM, based on the initial prompt, extracts relevant information from data. This includes the locations of existing base stations, areas with weak coverage, and data traffic. The prompt here serves not just as a question but as a complex instruction that enables the LLM to understand the context of the problem. \textbf{ii) Automatic Modeling \& Optimization}: By inputting well-designed prompts, the LLM will generate preliminary mathematical models for base station siting. These models include an objective function—such as minimizing deployment costs—as well as constraints, like minimum coverage requirements and minimum distance requirements between base stations. At this stage, the prompt must contain a detailed task description to ensure that the optimization objectives are correctly applied within the model. \textbf{iii) Automatic Code Generation \& Correction}: The LLM will generate code (e.g., Python code) to solve the model based on the input prompts. We then execute the code provided by the LLM to identify any errors. If there are bugs, error messages from the language environment (e.g., Python interpreter) are directly input as prompts into the LLM for correction. If the program runs successfully, we ultimately verify whether the BSS solution meets the constraints. If any discrepancies arise, this information is provided as a prompt to the LLM for further refinement.}

\begin{figure}[t]
\centerline{\includegraphics[width=0.45\textwidth, keepaspectratio]{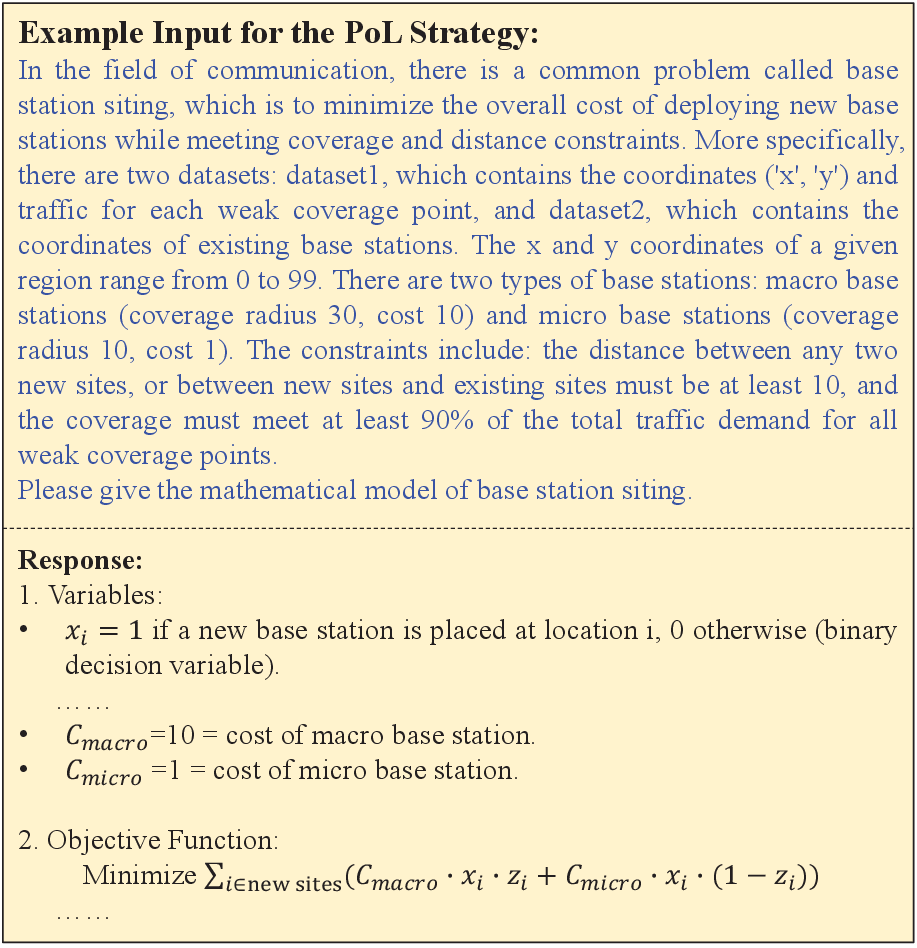}}
\caption{An example prompt for mathematical modeling of PoL strategy.}
\label{EPOL}
\end{figure}

{Fig. \ref{EPOL} presents a detailed example of the initial prompt and the generated preliminary response by the proposed PoL strategy.  
}

\subsubsection{Analysis of PoL Strategy}
{The PoL strategy exemplifies a Human-in-the-Loop workflow, characterized by the dynamic interaction between human users and the LLM to iteratively refine outputs for solving the BSS task. The primary roles of human involvement include:  i) Initial Prompt Design: Users craft an initial prompt to direct the LLM toward generating the desired output. If the generated results fall short of expectations, users revise and fine-tune the prompt based on the output to guide further iterations.  ii) Feedback and Refinement: LLM outputs are evaluated and validated by users, who provide feedback and make adjustments to the prompt as needed. This iterative cycle enables progressive optimization, gradually leading to improved results through repeated trials.   Moreover, when handling multi-faceted tasks or complex problems, human users are tasked with decomposing the overall objective into well-defined sub-tasks to facilitate effective problem-solving.  In summary, the PoL strategy is not fully automated. It heavily depends on human expertise for prompt design and optimization, making user oversight and intervention integral to its operation.}

\subsection{Strategies based on Agent Engineering} 
{
Prompt engineering offers a relatively simple and flexible approach but requires continuous human involvement. In contrast, agent engineering allows for automated task execution, offering scalability and self-learning capabilities that reduce the need for human intervention. However, it comes with higher development complexity and greater initial investment costs.
The key challenge of designing and deploying autonomous agent systems for the BSS problem lies in integrating task-oriented AI knowledge. This includes leveraging reinforcement learning, fostering collaboration within multi-agent systems, and tailoring optimization algorithms to the specific environment.  To accomplish BSS tasks with a high degree of autonomy and minimal human intervention, we further propose two sophisticated, fully intelligent LLM-driven frameworks: the LLM-empowered autonomous BSS agent (LaBa) and Cooperative multiple LLM-based autonomous BSS agents (CLaBa) strategies. We will now delve into the intricacies and merits of each approach.
}

\begin{figure*}[ht!]
\centerline{\includegraphics[width=0.85\textwidth, keepaspectratio]{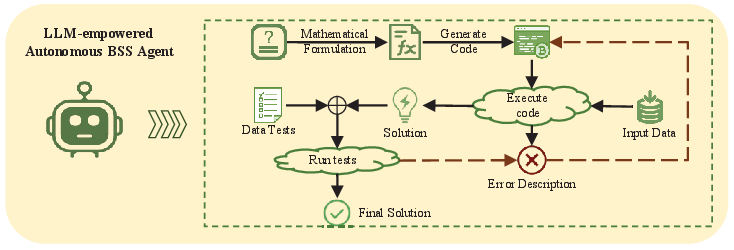}}
\caption{{Diagram of the LLM-empowered autonomous BSS agent (LaBa) strategy, illustrating the workflow from mathematical problem formulation and automated code generation to iterative debugging, error correction, and testing, ultimately delivering a validated solution for BSS task.}}
\label{fig4}
\end{figure*}

\subsubsection{Workflow of LaBa Strategy}

{As illustrated in Fig. \ref{fig4}, the workflow of the LaBa strategy comprises the following key steps, each representing a concrete application of LLM capabilities to BSS tasks:
\textbf{i) Problem Representation \& Modeling}: The user initially inputs details regarding the BSS task, which is typically articulated in natural language. Instead of directly passing all the information to the LLM, the relevant data is extracted from the task description and structured in a JSON file. This formatted file is then supplied to the LLM. Based on these inputs, the LLM formulates the optimization objectives (e.g., minimizing deployment costs) and constraints (e.g., minimum distance between base stations, minimum coverage requirements). The mathematical modeling step can be expressed as:
\begin{equation}
\arg\min\text{P}, s.t. \mathbf{C}\} \leftarrow \text{LLM}(x_0),
\end{equation}
where $P$ and $\mathbf{C}$ denote the objective function and the sets of constraint generated by by LLM, respectively, $x_0$ is the initial input of BSS problem description in JSON format.
\textbf{ii) Code Generation \& Execution}: Once the task objectives and constraints are analyzed, the LLM selects an appropriate optimization algorithm, such as Particle Swarm Optimization (PSO) \cite{PSO2022} or Genetic Algorithm (GA) \cite{sa2021}, and automatically generates the corresponding code. Alternatively, it may utilize existing optimization algorithm libraries like SciPy, Pyomo, or PuLP to address the BSS problem. The generated code, typically in Python or MATLAB, is executed within the simulation platform. There are two potential outcomes: either an execution error occurs, or the code runs successfully. In the event of an error, the error message is fed back to the LLM, which then revises the code to resolve the issue. This iterative debugging and modification process continues until the code executes without errors. \textbf{iii) Test \& Feedback-Driven Correction}: The LaBa strategy features real-time feedback and  multi-round iterative optimization capabilities. Simulation results are fed back to the LLM, and this feedback informs modifications to both the optimization model and the code. If the system identifies that the solution fails to meet the requirements (e.g., insufficient coverage), the LLM adjusts the optimization model and regenerates the code based on the feedback, iterating the process. This multi-round iterative feedback mechanism ensures that the resulting base station deployment scheme is well-suited to complex and dynamic real-world environments. The feedback adjustment process can be mathematically expressed as:
\begin{equation}
\mathcal{S}_{\text{new}} = \text{LLM}(\mathcal{S}, \mathcal{E}).
\end{equation}
where $\mathcal{S}_{\text{new}}$ and $\mathcal{S}$ denote the new and current solution, respectively, $\mathcal{E}$ represents the feedback detailing the issue, and the LLM uses this information to refine the deployment scheme.
}

\begin{figure*}[ht!]
\centerline{\includegraphics[width=0.85\textwidth, keepaspectratio]{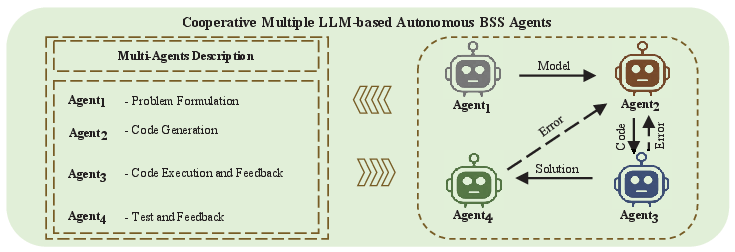}}
\caption{{Diagram of the cooperative multiple LLM-based autonomous BSS agents (CLaBa) framework, illustrating the workflow of task division and collaboration among agents: ${\rm Agent}_1$ for problem formulation, ${\rm Agent}_2$ for code generation, ${\rm Agent}_3$ for code execution and feedback, and ${\rm Agent}_4$ for testing and feedback, with iterative collaboration to refine and solve BSS task.}}
\label{fig5}
\end{figure*}

\subsubsection{Workflow of CLaBa Strategy}
{The CLaBa strategy distinguishes itself from the LaBa strategy by employing a collaborative multi-agent system. Rather than relying on a single agent to autonomously handle the entire BSS task, CLaBa distributes the task across multiple specialized agents. Each agent is responsible for a specific phase or subtask of the BSS optimization process, such as mathematical modeling, code generation, execution, and feedback-driven correction. As shown in Fig. \ref{fig5}, the workflow of CLaBa strategy consists of the following kep steps:
\begin{itemize}
\item \textbf{Problem Formulation \& Modeling (handled by ${\rm Agent}_1$)}: The CLaBa strategy begins with a dedicated agent, ${\rm Agent}_1$, formulating and modeling the problem. Similar to the LaBa strategy, the user articulates the problem in natural language, and relevant data (such as base station locations and traffic flow) are extracted and stored in a structured JSON format. This organization ensures that the input is systematically managed and easily processed by subsequent agents. The input data and task description are passed to ${\rm Agent}_1$, which generates the mathematical model defining the BSS optimization problem. The model includes objectives such as minimizing deployment costs, along with constraints like coverage requirements. This representation can be formalized as:
\begin{equation}
\arg\min\text{P},\quad s.t. \mathbf{C} \leftarrow {\rm Agent}_1(x_0),
\end{equation}
where $P$ and $\mathbf{C}$ denote the objective function and the sets of constraint generated by ${\rm Agent}_1$, respectively,  $x_0$ is the initial input of problem description in JSON format.
\item \textbf{Specialized Code Generation (handled by ${\rm Agent}_2$)}:
Once the mathematical model is established by ${\rm Agent}_1$, ${\rm Agent}_2$ is responsible for translating the model into executable code. The generated code accesses the necessary data directly from the JSON files created in the previous step. ${\rm Agent}_2$ focuses specifically on ensuring the efficiency and correctness of the code, making sure that the chosen optimization algorithm is correctly implemented. This clear division of responsibilities allows for a more streamlined development process, as each agent is optimized for its specific task.
\item \textbf{Execution of the Optimization Task (handled by ${\rm Agent}_3$)}:
After the code has been generated by ${\rm Agent}_2$, it is passed to ${\rm Agent}_3$, which is responsible for executing the code. ${\rm Agent}_3$ runs the optimization process using the provided data and code, and attempts to generate a solution for the BSS task. If any execution errors occur—such as issues with the optimization algorithm or data incompatibilities—${\rm Agent}_3$ passes the error message back to ${\rm Agent}_2$, which revises the code and attempts to resolve the issue. This feedback loop between ${\rm Agent}_2$ and ${\rm Agent}_3$ ensures that the code runs smoothly and produces valid outputs.
\item \textbf{Collaborative Testing \& Feedback-Driven Optimization (handled by ${\rm Agent}_4$)}: Once the optimization has been executed and a preliminary solution is generated, ${\rm Agent}_4$ takes charge of testing the solution. This agent uses test cases that are generated based on the initial task constraints (e.g., ensuring minimum coverage). Unlike the LaBa strategy, where a single agent handles all feedback and correction, CLaBa allows ${\rm Agent}_4$ to specialize in testing and validation. Users can also modify the test criteria at this stage to introduce domain-specific knowledge. If the solution fails to meet the test criteria (e.g., insufficient coverage), ${\rm Agent}_4$ sends feedback to ${\rm Agent}_2$, which modifies the code based on the identified issues.
\item \textbf{Iterative Feedback Loop \& Multi-Agent Collaboration}:
The key distinction of CLaBa lies in its iterative feedback loop across multiple agents. While LaBa relies on a single agent to handle all aspects of feedback and correction, CLaBa leverages the collaboration between ${\rm Agent}_2$, ${\rm Agent}_3$, and ${\rm Agent}_4$. This ensures that the generated solution undergoes continuous refinement, with each agent contributing its expertise to improve the solution. The feedback mechanism ensures that the process does not stop until an optimized, valid solution is found. The mathematical representation of this multi-agent feedback process is:
\begin{equation}
{\mathcal{S}}_{new} = {\rm Agent}_3 {\rm Agent}_2({\rm Agent}_4({\mathcal{S}}, {\rm \mathcal{E}})),
\end{equation}
where ${\rm Agent}_4$ identifies the errors, ${\rm Agent}_2$ corrects the code, and ${\rm Agent}_3$ re-executes the code to further refine the solution.
\end{itemize}
}

\begin{figure}[t]
\centerline{\includegraphics[width=0.45\textwidth, keepaspectratio]{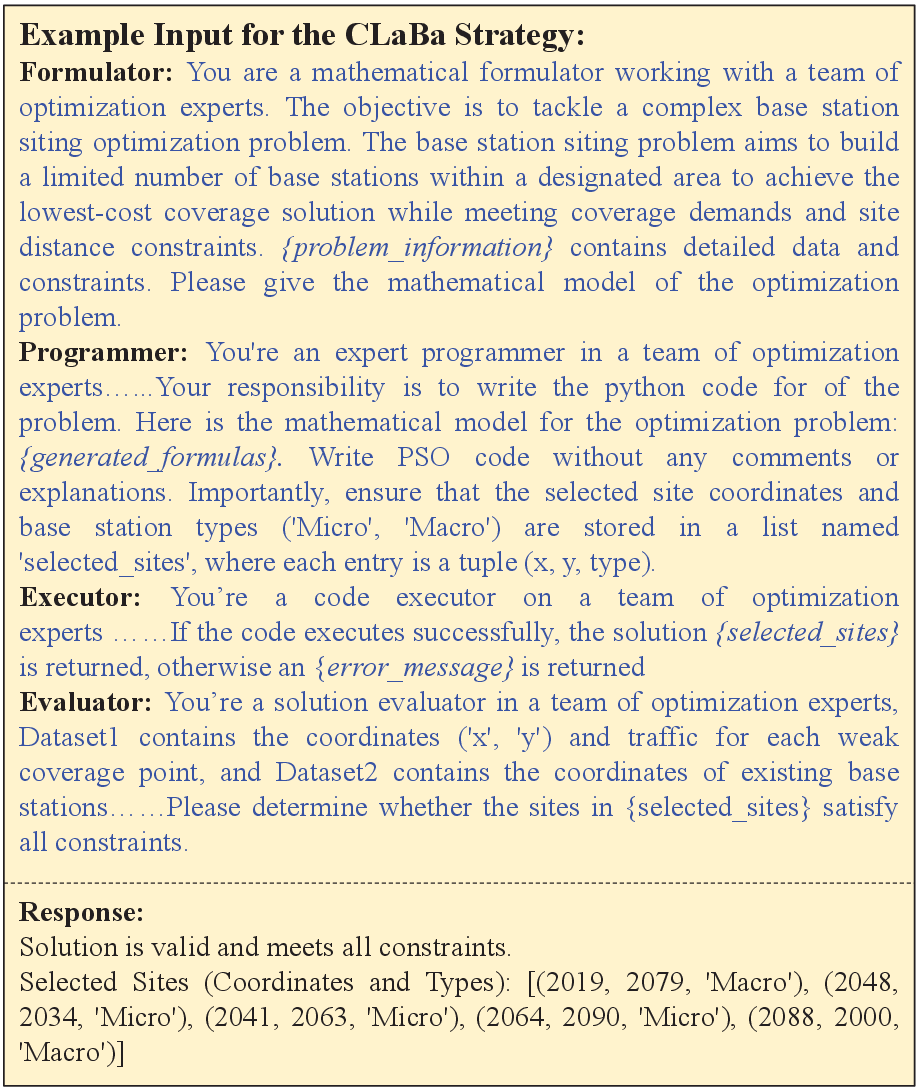}}
\caption{{Example prompts for mathematical modeling, programming, execution, and evaluation of CLaBa strategy.}}
\label{ECLABA}
\end{figure}

{Fig. \ref{ECLABA}  illustrates an example of the input for the CLaBa strategy, demonstrating the initialization of multiple agents for distinct sub-tasks and the corresponding generated solutions. For the sake of brevity, we do not present the prompts and generated response of the LaBa strategy here, as it is quite similar to the CLaBa approach. The main difference is that a single agent completes tasks such as modeling, programming, execution, and evaluation sequentially within the LaBa strategy, whereas the CLaBa strategy leverages multiple LLMs to perform different sub-tasks.}


\subsubsection{{Analysis of LaBa and CLaBa Strategies}}
{The LaBa strategy is straightforward to implement, with an intuitive architecture design and deployment process. One single agent employs a unified model for inference, producing clear, easily traceable results, which makes it efficient for solving one single well-defined problem. However, when tasks become more complex and require parallel processing, a single agent may become a bottleneck. Additionally, a single agent is vulnerable to single points of failure, as an error in any component in the workflow could lead to the overall task failing. In cases where tasks require knowledge or skills from multiple domains, a single agent may not offer sufficient coverage.}

{In contrast, the CLaBa strategy leverages multiple agents to simultaneously tackle different sub-tasks, significantly reducing overall task completion time. Each agent focuses on optimizing specific sub-tasks, enhancing both the accuracy and efficiency of the solution. Furthermore, when certain agents encounter errors, the remaining agents can continue working, thereby improving the system's fault tolerance. Multi-agent collaboration is particularly effective for addressing cross-domain or cross-module problems, especially when tasks involve multi-step reasoning or expertise in various fields.  This adaptability makes CLaBa suitable for evolving and dynamic network environments, where new constraints or objectives may arise. The cost of these performance gains is an increase in development complexity, as it requires the design of interaction protocols, collaboration strategies, and fault handling mechanisms between agents.}

\begin{figure}[ht!]
\centerline{\includegraphics[width=0.49\textwidth, keepaspectratio]{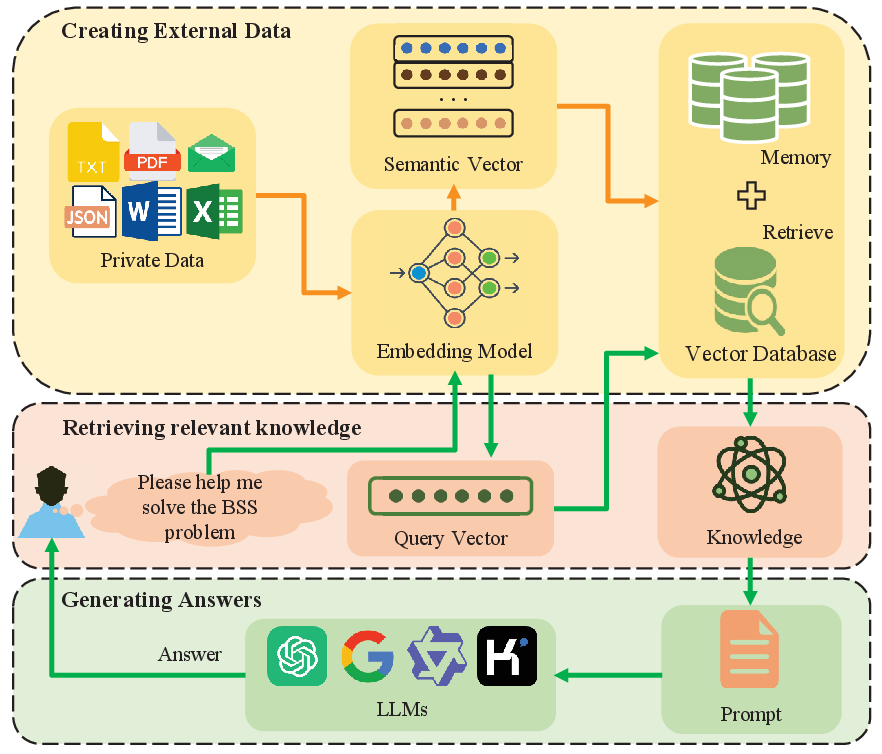}}
\caption{{Flowchart of RAG system. The system integrates private data, transforms it into semantic vectors using an embedding model, and stores these vectors in a vector database. User queries are converted into query vectors to retrieve relevant knowledge, which is then combined with an LLM to generate accurate responses.}}
\label{rag}
\end{figure}

\section{RAG-enhanced Strategy}
{
While LLMs equipped with prompt engineering and agent engineering have proven effective in NLP tasks, their reliance on pre-trained knowledge bases can limit their performance in domain-specific applications. The dynamic and complex nature of BSS demands up-to-date, contextually relevant information, which may not be readily available through pre-trained models. Therefore, there is a critical need for an approach that enhances LLMs with real-time, domain-specific knowledge to improve their decision-making capabilities in BSS optimization.
}

\subsection{Preliminary of RAG}
{
RAG offers a promising solution to these challenges by combining external knowledge retrieval with LLMs, effectively expanding the scope and depth of the model's knowledge base. By integrating information retrieval into the process of prompt generation and response creation, RAG enables LLMs to dynamically incorporate relevant external information, thereby enhancing the accuracy and relevance of the output. For instance, \cite{lewis2020retrieval} points out that RAG is particularly beneficial for tasks that require extensive background knowledge (e.g., open-domain question answering), as it allows for the real-time retrieval of latest  information. In \cite{10679152}, RAG was leveraged to support mathematical modeling and problem formulation in satellite communication networks. Similarly, in the context of BSS problems, RAG is well-suited to supplement real-time network data and domain-specific knowledge, improving the LLM’s ability to generate optimized BSS solutions.
}

\subsection{The Workflow of RAG-enhanced strategies for BSS task}
{
The implementation of RAG involves several key steps, as shown in Fig. \ref{rag}. First, a domain-specific knowledge base, referred to as external data, is created for the BSS application. This knowledge base may include real-time network performance data, optimization modeling methods, solution codes, and more, typically stored in various formats such as files, databases, or extended text. Next, an embedding model (e.g., text-embedding-ada-002 released by OpenAI\footnote{https://zilliz.com/ai-models/text-embedding-ada-002}) is employed to transform this data into vector representations, which are then stored in a vector database. This process results in a dynamic knowledge base that can be accessed by the LLM, enriching its ability to generate accurate and contextually relevant solutions.
}

{Following this, the system performs relevance matching between the user query and the data in the knowledge base. The user query $q$ is first transformed into a vector representation \(\mathbf{q}\) using the embedding model. Then, cosine similarity is calculated between \(\mathbf{q}\) and each vector \(\mathbf{v}_i\) in the database to obtain the correlation score:
\begin{equation}
{\rm sim}(\mathbf{q},\mathbf{v}_i) = \frac{\mathbf{q} \cdot \mathbf{v}_i}{||\mathbf{q}||||\mathbf{v}_i||}.
\end{equation}
Based on these scores, the system selects the top \(k\) most relevant data items, denoted as \(d_{i_1}, d_{i_2}, \dots, d_{i_k}\), to serve as supplementary information for the LLM. For example, in a BSS task where the prompt is "How to deploy base stations in weak coverage areas to improve signal quality," the system will retrieve relevant content related to optimization modeling and solution code, assisting the LLM in generating accurate optimization recommendations.
}

{In the final step, RAG enhances the LLM's response by incorporating the retrieved relevant data into the prompt. The updated prompt includes both the user's original query and the retrieved knowledge base information \(\{q, d_{i_1}, d_{i_2}, \dots, d_{i_k}\}\), enabling the LLM to better understand the task requirements and generate a deployment plan that meets the BSS criteria. In generating the final response, the LLM integrates the latest BSS-related information with its original training data.
}

\section{Experimental Results}

{In this work, we propose three innovative strategies that leverage the capabilities of LLMs, marking a significant paradigm shift in addressing the BSS problem. The first strategy centers on prompt engineering, emphasizing the dynamic interaction between human operators and LLMs to foster a collaborative problem-solving approach. In contrast, the latter two strategies focus on autonomy, advocating for end-to-end automated solutions driven by LLM-empowered AI agents. These strategies not only expand the problem-solving landscape for BSS but also provide a comprehensive framework to evaluate the merits of both human-LLM collaboration and AI-driven automation.

It is worth noting that LLM-based methods complement, rather than compete with, traditional approaches. When integrated with conventional methods, LLMs can serve as an augmentative tool, enhancing their effectiveness. Specifically, LLMs can address particular challenges by utilizing open-source algorithm toolkits or expert knowledge libraries curated by engineers. The prompt engineering-based strategy allows engineers to guide LLMs in selecting specific algorithms, such as genetic algorithms or deep reinforcement learning, based on the performance of generated solutions. This approach offers greater flexibility and adaptability compared to traditional methods. On the other hand, strategies based on agent engineering rely entirely on the autonomous learning processes of agents to determine which algorithms to invoke. This level of autonomy presents a novel solution to the BSS problem, potentially outperforming traditional methods in specific scenarios. The following subsections provide detailed experimental results to validate the effectiveness and reliability of the proposed strategies.
}

\subsection{Experimental Setup}
{\subsubsection{Dataset}
In this study, we utilize a dataset that reflects real-world conditions for mobile communication network site planning in urban scenario\footnote{http://www.mathorcup.org}. This dataset provides a comprehensive view of the coverage delivered by existing base stations in urban settings, as well as identifies regions experiencing suboptimal signal strength. Meticulously divided into a grid of $2500 \times 2500$ units on an authentic map, the dataset furnishes granular network and traffic data for the centroid of each grid cell. This encompasses accurate geographic coordinates, traffic volume, and flags indicating weak coverage areas. Additionally, the dataset encompasses the geographical coordinates of existing base stations, an essential element for strategic planning and optimization tasks.

\subsubsection{System Parameters}
For this study, we define the coverage radii for macro and micro base stations as $d_h=30$ grids and $d_d=10$ grids, respectively. The associated deployment costs are set as $C_h=10$ for macro base stations and $C_d=1$ for micro base stations. To ensure network integrity and minimize interference, we impose a minimum separation distance of $D_{\min}=10$ grids between any two base stations. The primary objective of this study is to enhance network coverage in previously underserved areas while simultaneously minimizing deployment costs. Specifically, we aim to extend coverage to at least $\theta_{cp}=90\%$ of the total traffic volume in these regions. The choice of the $90\%$ threshold reflects a balanced trade-off between practical feasibility and ambitious optimization goals. This coverage target not only simulates real-world challenges, but also ensures that the proposed methods can effectively support key traffic areas, while also controlling costs.

\subsubsection{Baselines}
To validate the effectiveness of our proposed LLM-based methods, we compare them against two widely-used traditional approaches in BSS task:
\begin{itemize}
\item \textbf{Particle Swarm Optimization (PSO) Method \cite{PSO2022}}: A well-established metaheuristic optimization algorithm, PSO has been extensively applied in various network planning problems. It optimizes solutions through iterative improvement based on a predefined quality measure, making it an ideal benchmark for comparison in this study.
\item \textbf{Simulated Annealing (SA) Method \cite{sa2021}}: SA is a probabilistic optimization technique inspired by the physical process of annealing. By allowing the acceptance of worse solutions with a certain probability, SA explores the solution space more comprehensively, enabling it to escape local optima and approach the global optimum. Its capacity to navigate complex solution landscapes makes it a valuable comparison method for this study.
\end{itemize}
For a fair comparison, both baseline methods are implemented under identical experimental conditions, including the same dataset and defined constraints.

\begin{figure}[t]
\centerline{\includegraphics[width=0.45\textwidth, keepaspectratio]{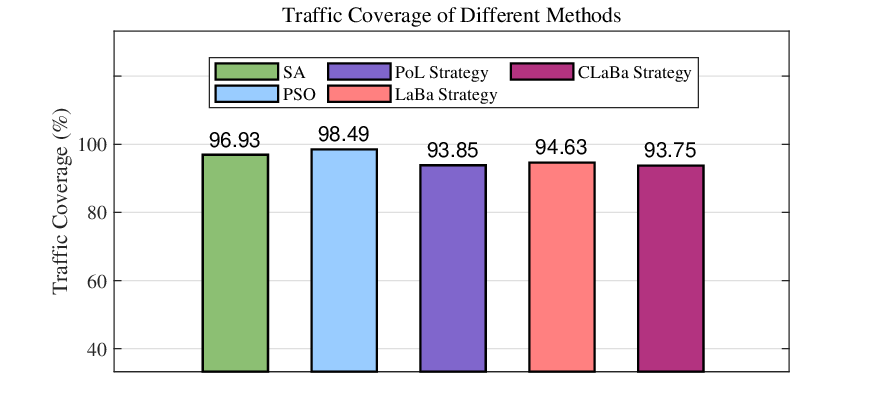}}
\caption{{Traffic coverage comparison across different methods.}}
\label{TR}
\end{figure}

\begin{figure}[t]
\centerline{\includegraphics[width=0.45\textwidth, keepaspectratio]{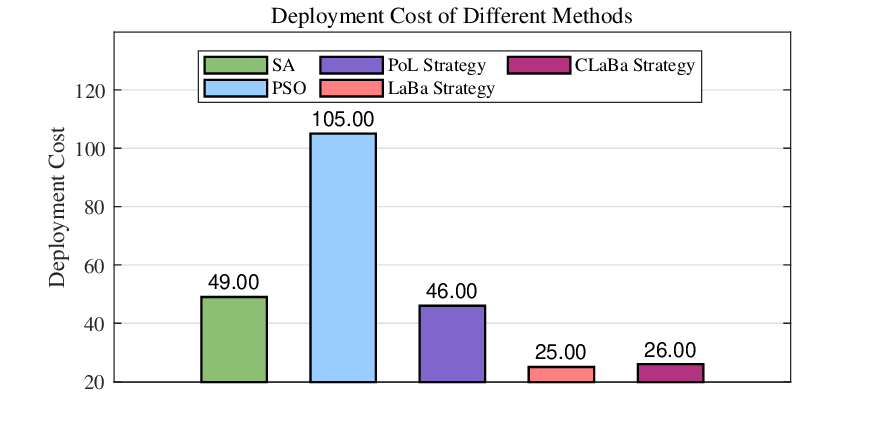}}
\caption{{Deployment cost comparison across different methods.}}
\label{CT}
\end{figure}

\subsubsection{Metrics}
In this work, we utilize four key metrics: traffic coverage, deployment cost, success rate, and execution time. These metrics offer a comprehensive evaluation of each strategy's performance across different dimensions.
\begin{itemize}
\item \textbf{Traffic Coverage:} This is a crucial factor in BSS, as it directly impacts both network performance and user satisfaction. It is defined as the proportion of total traffic within a given area that is covered by base stations. Achieving high traffic coverage is essential for maintaining service quality and avoiding network congestion. In this study, we aim for at least $\theta_{cp} = 90\%$ traffic coverage, ensuring that the network operates efficiently even under high traffic volumes.
\item \textbf{Deployment Cost:} This metric represents the total cost of deploying the necessary base stations to achieve the target traffic coverage. It is vital for telecommunications operators to ensure the economic feasibility of network expansion by optimizing performance within budget constraints. 
\item \textbf{Success Rate:} The success rate refers to the proportion of effective solutions generated by each strategy that meet all predefined constraints, such as coverage, minimum distance, and traffic demand. This metric evaluates the reliability and robustness of each strategy. A higher success rate indicates that the strategy consistently produces feasible solutions, which is critical for the long-term success of network deployments. 
\item \textbf{Execution Time:} This metric measures the total time required for each strategy to generate the final solution, including data processing, model generation, execution, and feedback adjustments. It assesses the computational efficiency of each strategy, especially for the more automated LaBa and CLaBa approaches. Shorter execution times suggest that a strategy can quickly adapt to the dynamic needs of network planning. 
\end{itemize}
}

\begin{figure*}[ht!]
\centerline{\includegraphics[width=1\textwidth, keepaspectratio]{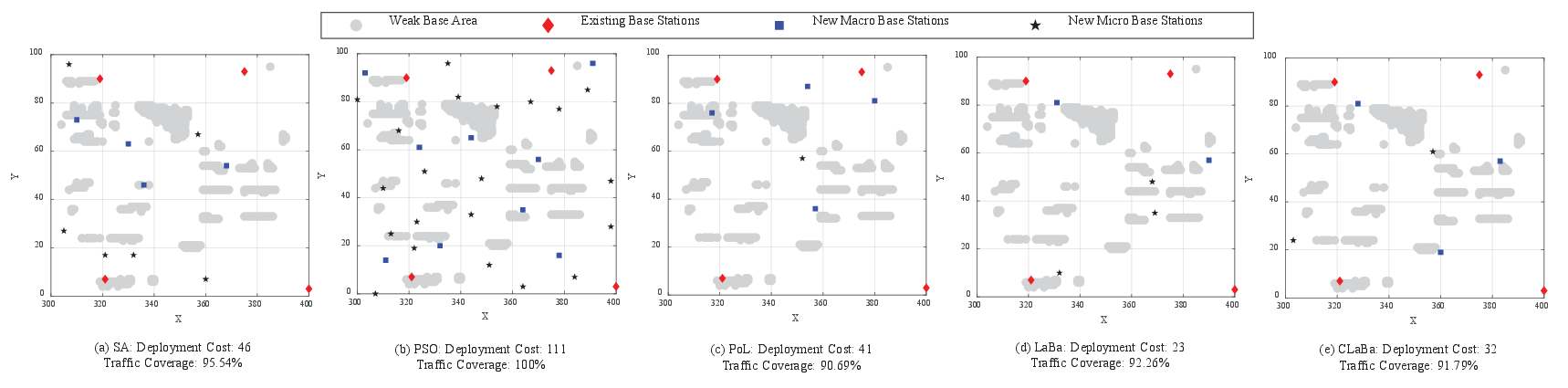}}
\caption{Visualized results of the BSS solutions generated by different methods, including weak coverage areas, existing base stations, new macro base stations, and new micro base stations.}
\label{ZH}
\end{figure*}

\subsection{Experiment Results}
{In this subsection, we present a detailed analysis of the experimental results, which substantiate the effectiveness of the proposed LLM-based BSS optimization strategies. To balance computational efficiency with experimental representativeness, we randomly selected 25 distinct $100 \times 100$ regions from the dataset, instead of using the entire $2500 \times 2500$ grid. By averaging the results across these 25 regions, we ensure that the performance of the proposed strategies is assessed under diverse network conditions.

\subsubsection{Performance Comparison}
As depicted in Fig.~\ref{TR}, the average traffic coverage across 25 randomly selected regions achieved by the proposed LLM-based strategies is compared against that of the baseline methods. The PSO method achieves the highest coverage rate of   $98.49\%$, while the PoL, LaBa, and CLaBa strategies also fulfill the $90\%$ coverage requirement with strong performances of $93.85\%$, $94.63\%$, and $93.75\%$, respectively. The SA method attains a coverage rate of $96.93\%$. These results indicate that, despite not surpassing the PSO method's high coverage, the proposed LLM-based strategies still successfully meet the $90\%$ coverage target and exhibit significant advantages in cost-effectiveness and computational efficiency, as further demonstrated in Fig.~\ref{CT}.

Fig.~\ref{CT} compares the average deployment costs across 25 randomly selected regions of these methods. The PoL strategy achieves $93.85\%$ traffic coverage at a cost of $46$, while the LaBa strategy provides $94.63\%$ coverage at a lower cost of $25$. The CLaBa strategy offers $93.75\%$ coverage at a cost of $26$. In contrast, the PSO method incurs the highest deployment cost of 105, while the SA method also has a relatively high cost of 49. Together with the results in Fig.~\ref{TR}, these findings underscore the clear cost advantage of the proposed LLM-based strategies, highlighting their ability to effectively control costs while still meeting the $90\%$ traffic coverage requirement.

To provide a more intuitive understanding of the effectiveness of the proposed strategies, Fig.~\ref{ZH} visualizes the outcomes of different approaches to solving the BSS problem within one randomly selected region. The figures show weak coverage areas, existing base stations, and the newly selected macro and micro base stations. The proposed LLM-based strategies feature a well-balanced layout of macro and micro base stations, ensuring effective coverage of weak areas while efficiently managing costs.

\begin{figure}[t]
\centerline{\includegraphics[width=0.45\textwidth, keepaspectratio]{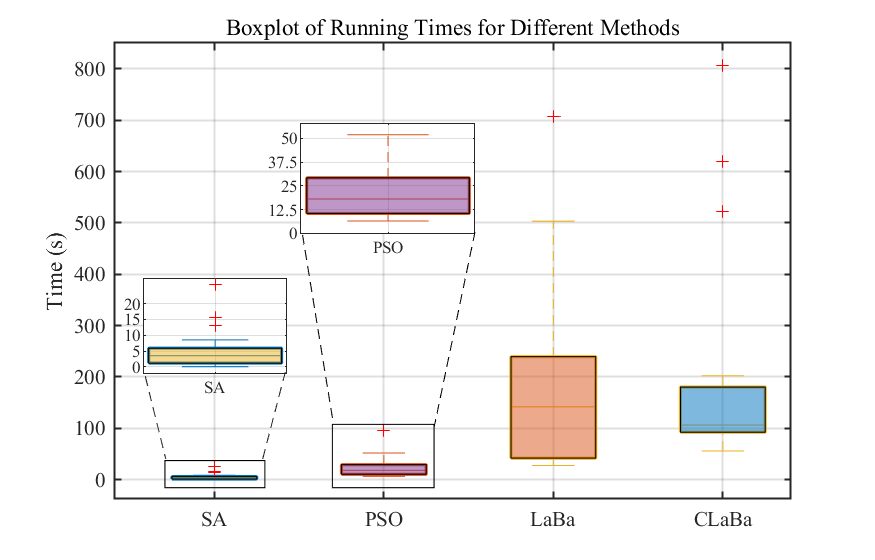}}
\caption{{Computational efficiency of different methods in solving BSS tasks.}}
\label{time}
\end{figure}

To demonstrate the computational efficiency of our proposed method, Fig.~\ref{time} presents the running time of different methods in solving the BSS task. Notably, the running time for the PoL strategy is not provided, as it involves human interaction, which introduces variability and makes accurate time measurement challenging. As shown in the figure, traditional methods, specifically SA and PSO, exhibit higher computational efficiency, requiring less execution time compared to the LLM-based strategies. In contrast, the proposed LLM-based strategies, namely LaBa and CLaBa, take longer to complete due to their iterative feedback mechanisms and complex decision-making processes. Furthermore, it is worth mentioning that the execution times for PSO and SA here only reflect the algorithm's running time, excluding the time spent by human engineers on tasks such as data collection, data analysis, model development, and algorithm selection.   Although these methods have higher computational times compared to traditional approaches, they offer more nuanced optimization solutions, as indicated by the results in Fig.~\ref{TR} and Fig.~\ref{CT}.  The LLM-based strategies provide a better balance between coverage and cost. Additionally, the LLM-based strategies eliminate the need for human intervention, achieving full automation in the decision-making process.

\begin{table}[]
\caption{Comparison of LLM-based and Traditional Methods}
\centering
\renewcommand\arraystretch{1.3}
\footnotesize
\label{table}
\begin{tabular}{|c|l|l|}
\hline
& \multicolumn{1}{c|}{\begin{tabular}[c]{@{}c@{}}\textbf{LLM-based Methods}\end{tabular}}                       & \multicolumn{1}{c|}{\begin{tabular}[c]{@{}c@{}}\textbf{Traditional Methods}\end{tabular}}                            \\ \hline
\textbf{Efficiency}  &  Automated & Labor-dependent 
\\ \hline
\textbf{Deployment} & High initial investment  & Low initial cost \\ \hline
\textbf{Maintenance}  & Low long-term cost & High long-term cost
\\ \hline
\textbf{Flexibility}                                         & \begin{tabular}[c]{@{}l@{}}Strong, easily updatable\\ and adjustable \end{tabular}             & \begin{tabular}[c]{@{}l@{}}Weak, high cost to\\ update and adjust\end{tabular}                 \\ \hline
\textbf{Scalability }  & \begin{tabular}[c]{@{}l@{}} Scalable for other tasks \\ and networks \end{tabular}          & \begin{tabular}[c]{@{}l@{}} Fixed paradigms with \\ limited scalability \end{tabular}  \\ \hline
\begin{tabular}[c]{@{}c@{}}\textbf{Human}\\ \textbf{Intervention}\end{tabular}    & \begin{tabular}[c]{@{}l@{}}Minimal manual \\  intervention\end{tabular}                        & \begin{tabular}[c]{@{}l@{}}Highly dependent on\\ manual decision-making\\ and feedback\end{tabular}           \\ \hline
\textbf{Real-time }                                                     & \begin{tabular}[c]{@{}l@{}}Real-time data processing \\ with quick response\end{tabular}       & \begin{tabular}[c]{@{}l@{}}Limited real-time\\ capability, long update \\ cycles\end{tabular}                    \\ \hline
\begin{tabular}[c]{@{}c@{}}\textbf{Technical}\\ \textbf{Dependency}\end{tabular} & \begin{tabular}[c]{@{}l@{}}Rely on data and \\ computing devices \end{tabular} & \begin{tabular}[c]{@{}l@{}}Rely on data and \\ experts' experience \end{tabular} \\ \hline
\end{tabular}
\end{table}

In summary, the advantages and disadvantages are listed in Table. \ref{table} to compare the proposed LLM-based methods and traditional methods.

\begin{figure}[t]
\centerline{\includegraphics[width=0.45\textwidth, keepaspectratio]{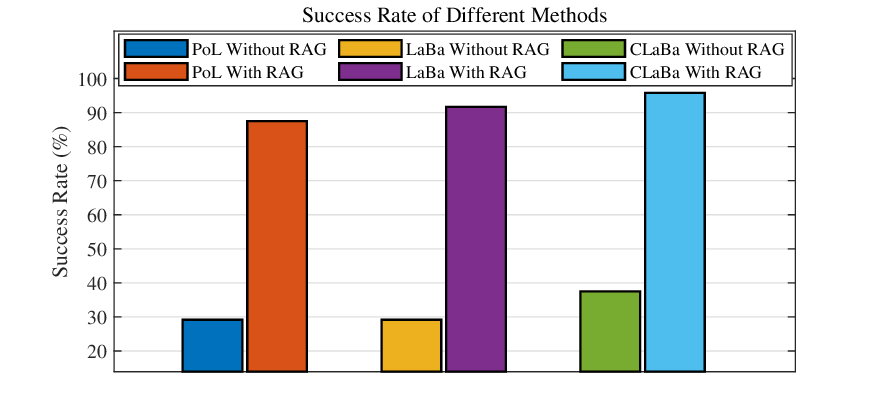}}
\caption{{Success rate comparison between the proposed methods with and without RAG.}}
\label{SR}
\end{figure}

\begin{figure}[t]
\centerline{\includegraphics[width=0.45\textwidth, keepaspectratio]{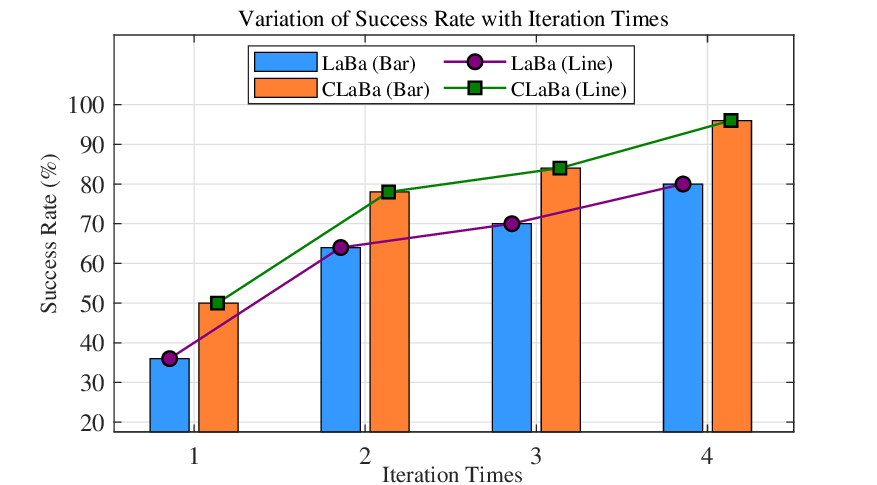}}
\caption{{Success rate variation of LaBa and CLaBa strategies with iteration times.}}
\label{srit}
\end{figure}

\subsubsection{Ablation Study}
To verify the effectiveness and reliability of the proposed LLM-based strategies, we evaluate their performance based on the success rate. In the PoL strategy, we assume a maximum of 10 interactions between LLM and humans. Similarly, for the LaBa and CLaBa strategies, the maximum number of iterative optimizations is set to 10. If the number of interactions or iterations exceeds these limits, the strategy is considered a failure. This approach reflects the need for decision-makers to make timely and effective decisions within a limited timeframe, which is typical in real-world network deployment scenarios. We conduct BSS optimization by randomly selecting 25 areas, and the proportion of successful deployments in these areas represents the success rate.

Fig.~\ref{SR} presents the success rates of different strategies, with particular emphasis on the use of the RAG technique. When the proposed strategies are not integrated with RAG, their success rates are relatively low. However, among these, the CLaBa strategy achieves the highest success rate. This is because each agent in CLaBa specializes in a specific task, such as mathematical modeling, code generation, and solution validation, thereby improving performance in those areas. After incorporating RAG, the success rates of the proposed strategies increase significantly, reaching over $80\%$, nearly doubling compared to when RAG is not used. This highlights that RAG, by providing domain-specific knowledge, greatly enhances the accuracy and robustness of the strategies.

Exploring the impact of iteration limits on the success rate is essential for evaluating the efficiency of the strategies. In real-world applications, solutions are often constrained by time and computational resources. Therefore, examining how different iteration limits affect the success rate helps decision-makers strike a balance between efficiency and resource consumption, optimizing performance and robustness under limited conditions. 

Figure~\ref{srit} illustrates how the success rates of the LaBa and CLaBa strategies change with different iteration limits. he results indicate that when the iteration limit is 1, both strategies have relatively low success rates. As the iteration limit increases, the success rates of both strategies improve significantly, suggesting that the agents can learn and adapt more effectively through additional attempts, thereby enhancing their success rates. Furthermore, the success rate of the LaBa strategy is consistently lower than that of the CLaBa strategy, demonstrating the superiority of the multi-agent framework in solving complex tasks by dividing and collaborating on different subtasks.
}

\section{Discussions}
In this section, we explore several open issues and also promising directions for future research and development in the integration of LLMs with next-generation networks and communications.

{
\subsection{Solution for Addressing Potential Limitations of LLMs}
To effectively implement LLMs in practical applications, it is crucial to address their limitations, such as dependency on data quality and the need for regular updates. LLMs, like all data-driven models, are significantly influenced by the quality and relevance of the data used for training. This challenge can be mitigated by incorporating techniques like real-time data retrieval (e.g., via RAG-based approaches), which allows the model to access up-to-date, domain-specific information as required. This enables the LLM to adapt dynamically to changes in data, thereby improving both its robustness and accuracy. Additionally, the model’s performance may degrade if the data it was originally trained on becomes outdated. To ensure flexibility and scalability, techniques such as \textbf{F}ine-tuning, \textbf{I}n-context learning, \textbf{C}hain-of-thought reasoning, \textbf{T}ool-calling, \textbf{RA}G, and \textbf{M}ulti-agent (FICTRAM) techniques can be employed to seamlessly integrate the latest updates without the need for full model re-training. This approach further enhances the model's capability to reason through evolving and dynamic information.}

{
\subsection{Enhancing Framework Applicability through Open-Source and Localized LLM Deployment}
In our studies, the use of closed-source LLMs, such as GPT, has provided strong support for the validation of our framework. However, this cloud service-dependent model may face applicability challenges in scenarios without network connectivity. Open-source LLMs, such as LLaMA, OPT, or Bloom, offer a practical solution to this problem. Open-source models can not only operate in environments without network connectivity through localized deployment but can also be fine-tuned for specific scenarios, thereby enhancing their adaptability. Furthermore, the integration of model optimization techniques, such as quantization and pruning, can further reduce the model's dependence on hardware resources, enabling it to run efficiently on devices with limited computational and storage capabilities. This direction of improvement will significantly enhance the flexibility and universality of the framework.}

\subsection{LLM-empowered AI Native Next-Generation Networks}
The native intelligence of the next generation communication network can be rapidly established and boosted by fully utilizing LLM's potent natural language processing capability, the native intelligence of the upcoming generation of communication networks. For example, in future communication and networks, resource management is a core task to ensure efficient network operation. LLMs and other AI technologies play a significant role in resource management by improving the utilization efficiency of network resources through intelligent scheduling and optimization. Specifically, LLMs can analyze historical data and current network status, predict future network needs, optimize resource allocation in advance, and reduce network congestion and latency. The integration of LLMs does, however, come with certain difficulties, including designing flexible interfaces to adapt to different network environments, developing efficient algorithms to meet real-time requirements, and optimizing models to accommodate the resource constraints of network devices. By adopting a modular design, different components of the LLMs can be integrated into the network system as needed. Algorithm optimization can reduce computing resource consumption to ensure fast responses. Additionally, flexible interface design ensures that LLMs can operate efficiently in various network environments.

\subsection{Task-oriented Selection in Human-LLM Interaction or Autonomous Agents}
For the future generation of networking and communication systems, it is essential to make the task-oriented decision between a human-LLM interaction framework or a fully automated LLM framework. On the one hand, the purely autonomous LLM-based framework can significantly improves efficiency by reducing human involvement. However, LLMs are known to suffer from the hallucination problem, where models can produce inaccurate or misleading information. This issue is particularly severe in automated network management and communication systems, where it can result in hazards and faults in the system. On the other hand, human-LLM interaction can mitigate the impact of hallucinations, improving system reliability. Human involvement can serve as a verification and correction mechanism to detect and correct erroneous information generated by LLMs promptly. For example, in an automated customer service system, customer service personnel can review and adjust the model's responses to ensure users receive accurate and reliable service. Although this approach may reduce overall efficiency, it enhances the accuracy and reliability of information, increasing user trust and reducing potential risks.

\section{Conclusion}
This study explored the potential of LLMs in optimizing BSS problem and proposed three innovative strategies: PoL, LaBa, and CLaBa. Each strategy demonstrated distinct advantages, ranging from reducing human intervention to enabling highly automated and adaptive solutions. Experimental evaluations showed that the proposed methods effectively balanced traffic coverage, and deployment cost, thus meeting the requirements of real-world scenarios. Moreover, the integration of LLMs with RAG significantly improved the accuracy and robustness of the solutions, providing a solid foundation for solving complex optimization problems.

Future research is expected to build upon this framework and explore broader applications of LLMs in communication systems, such as dynamic resource management and intelligent decision-making. By combining human expertise with AI capabilities, the proposed framework paves the way for fully autonomous and scalable solutions, advancing the evolution of AI-driven engineering practices.

\bibliographystyle{IEEEtran} 
\bibliography{IEEEabrv,references}
\end{document}